\documentclass[conference]{IEEEtran}
\IEEEoverridecommandlockouts

\usepackage{cite}
\usepackage{amsmath,amssymb,amsfonts}
\usepackage{algorithmic}
\usepackage{graphicx}
\usepackage{textcomp}
\usepackage{xcolor}
\usepackage{booktabs}
\def\BibTeX{{\rm B\kern-.05em{\sc i\kern-.025em b}\kern-.08em
    T\kern-.1667em\lower.7ex\hbox{E}\kern-.125emX}}

\begin{document}

\title{%
An End-to-End DNN Inference Framework \\for the SpiNNaker2 Neuromorphic MPSoC

\thanks{Partially funded by the German Research Foundation (DFG, Deutsche Forschungsgemeinschaft) as part of Germany’s Excellence Strategy – EXC 2050/1 – Project ID 390696704 – Cluster of Excellence “Centre for Tactile Internet with Human-in-the-Loop” (CeTI) of TU Dresden, by BMBF and the Free State of Saxony within the ScaDS.AI center of excellence for AI research, as well as by BMBF within DAAD project SECAI (project ID: 5761681) and within joint project ”EVENTS” (funding reference: 16ME0729K).}
}

\author{
    \IEEEauthorblockN{Matthias Jobst\IEEEauthorrefmark{1}\IEEEauthorrefmark{2}, Tim Langer\IEEEauthorrefmark{1}, Chen Liu\IEEEauthorrefmark{1}\IEEEauthorrefmark{3}, Mehmet Alici\IEEEauthorrefmark{3}, Hector A. Gonzalez\IEEEauthorrefmark{3}, Christian Mayr\IEEEauthorrefmark{1}\IEEEauthorrefmark{2}}
    \IEEEauthorblockA{\IEEEauthorrefmark{1}Chair of Highly-Parallel VLSI Systems and Neuro-Microelectronics, TU Dresden, Germany}
    \IEEEauthorblockA{\IEEEauthorrefmark{2}Centre for Tactile Internet with Human-in-the-Loop (CeTI), TU Dresden, Germany}
    \IEEEauthorblockA{\IEEEauthorrefmark{3}SpiNNcloud Systems GmbH, Dresden, Germany}
    \IEEEauthorblockA{\{matthias.jobst2, tim\_hauke.langer, christian.mayr\}@tu-dresden.de\\\{chen.liu, mehmet.alici, hector.gonzalez\}@spinncloud.com}
}

\maketitle

\begin{abstract}

This work presents a multi-layer DNN scheduling framework as an extension of \textit{OctopuScheduler}, providing an end-to-end flow from PyTorch models to inference on a single SpiNNaker2 chip. Together with a front-end comprised of quantization and lowering steps, the proposed framework enables the edge-based execution of large and complex DNNs up to transformer scale using the neuromorphic platform SpiNNaker2.

\end{abstract}


\section{Introduction}

The efficient deployment of Deep Neural Networks (DNNs) on constrained devices has the potential to revolutionize the entire edge industry. While the primary energy challenges are associated with datacenter workloads \cite{shankar2024challenging}, mapping DNN models efficiently to the edge enables the development of smarter infrastructure nodes. Neuromorphic computing stands out as a particularly promising approach to significantly reduce the energy footprint of these AI workloads by emulating the extreme efficiencies of biological brains \cite{kudithipudi2025neuromorphic}.

\begin{figure}
    \centering
    \includegraphics[width=0.9\linewidth]{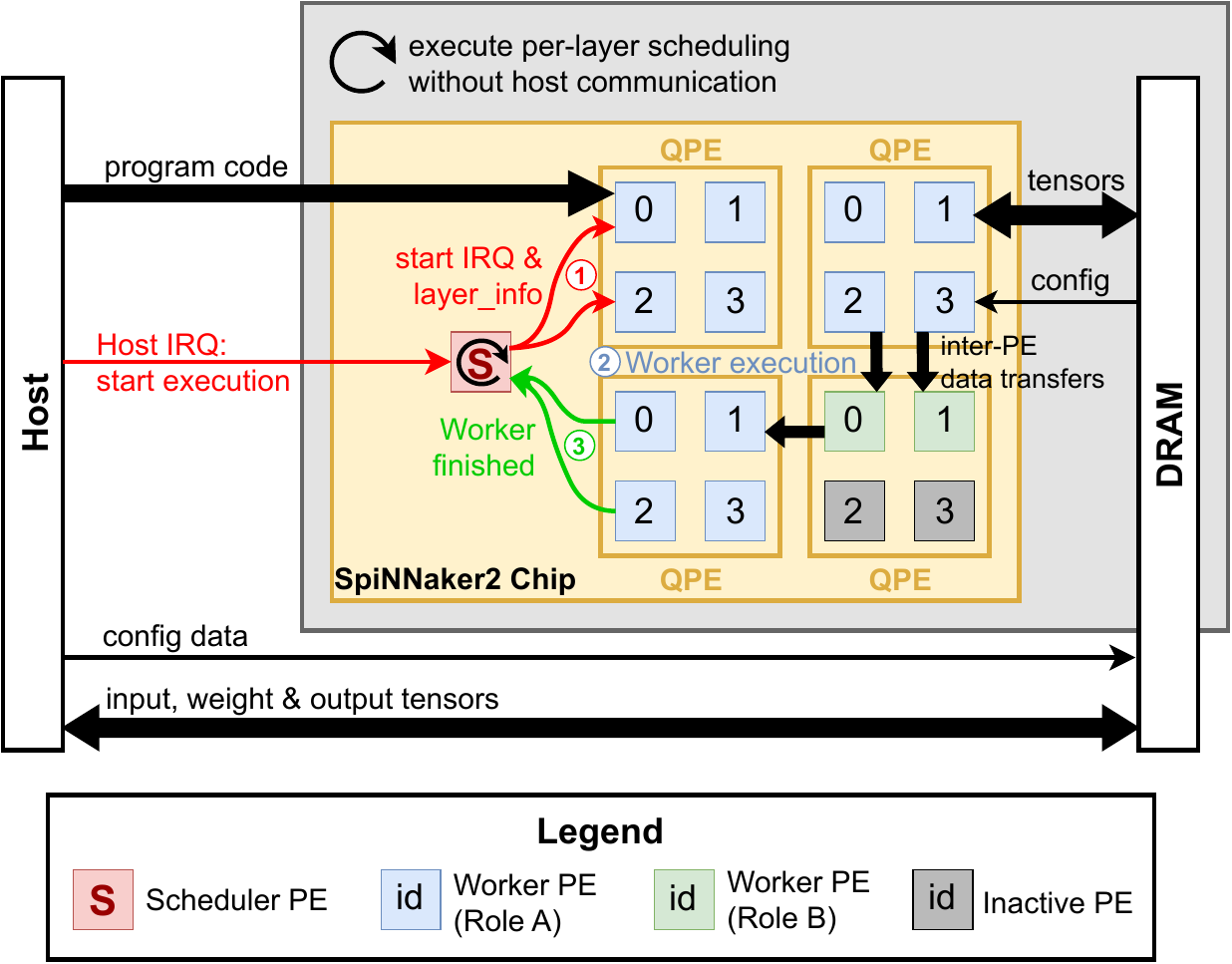}
    \caption{Overview of execution flow within the multi-layer extension of OctopuScheduler: After initial configuration from host, the SpiNNaker2 chip iterates over all layers of a model.}
    \label{fig:octopuscheduler_overview}
\end{figure}

SpiNNaker2 \cite{gonzalez_spinnaker2_2024} is a strategic neuromorphic multi-processor system-on-chip (MPSoC) designed to deploy large-scale spiking neural networks (SNN) \cite{arfa_efficient_2025, gerhards_hybrid_2023} deep neural networks (DNN) \cite{kelber_mapping_2020}, as well as hybrid event-based approaches \cite{nazeer_language_2024, bena2024eventbased}. SpiNNaker2-based systems are built both at the edge and supercomputer levels. In contrast to standard building blocks in HPC systems, a SpiNNaker2 chip consists of 152 low-power processing elements (PEs), each containing an ARM Cortex M4F processor as well as dedicated hardware accelerators for exponential functions (EXP) or machine learning (MLA). The distributed PEs each have a local 128\,kB SRAM and are connected via a network-on-chip (NoC) in a Globally Asynchronous Locally Synchronous (GALS) topology, allowing inter-PE transfers via direct memory access (DMA) and synchronization or control notifications via interrupt requests (IRQ). Every SpiNNaker2 chip is paired with an external 2\,GB DRAM memory to extend on-chip storage as required for large complex DNN models.

Although SpiNNaker2 can deploy large DNN networks across its numerous PEs within large-scale supercomputers, scheduling complex DNNs on single-chip edge setups is not a trivial task, as most models require multiple iterations through the DRAM and PEs. 
To implement DNNs on an MPSoC with low memory per PE, the individual DNN layers must be tiled and mapped to the single PEs. While more advanced partitioning and mapping frameworks such as GSPMD \cite{xu_gspmd_2021}, LOMA \cite{symons_loma_2021} or DeFiNES \cite{mei2024defines} exist, the \textit{OctopuScheduler} framework \cite{octopuscheduler25} was developed specifically for the SpiNNaker2 architecture utilizing existing platform-specific low-level libraries and high-level frameworks such as PySpiNNaker2 \cite{vogginger2024pyspinnaker2} as an initial approach for tiling, mapping and scheduling DNNs on SpiNNaker2.\\
This paper extends the single-layer on-chip DNN scheduling framework \textit{OctopuScheduler} \cite{octopuscheduler25} for SpiNNaker2 at the edge, with the following main contributions:
\begin{enumerate}
    \item  Extension of \textit{OctopuScheduler} towards multi-layer scheduling, paving the way for the edge inference of large complex state-of-the-art DNN models (e.g. transformers) on SpiNNaker2.
    \item  Standalone on-chip iteration over all DNN layers without inter-layer host communication.   
    \item  Automated flow for parameter extraction and 8-bit post-training quantization (PTQ) from PyTorch models.
\end{enumerate}

The paper is organized as follows. Section II presents the single-layer scheduling approach that this work expands upon. Section III outlines the extended multi-layer scheduling, while Section IV describes the full flow. Section V tests it using a dummy model, and Section VI situates this work within the broader SpiNNaker2 software ecosystem.

\section{OctopuScheduler: Single-Layer Scheduling on SpiNNaker2}
\textit{OctopuScheduler} is a framework for mapping and executing DNN layers on a SpiNNaker2 chip. 
DNN layers are split into equally sized tiles to allow for execution on multiple \textit{worker PEs} in a distributed fashion, taking into account the limited amount of PE-local memory with an automated design space exploration. A single \textit{scheduler PE} oversees and synchronizes all \textit{workers}. At the start of a layer, the \textit{scheduler} triggers the \textit{workers} to execute the layer-specific functions via an interrupt request (IRQ). Upon completion, the \textit{workers} send a signal word to the \textit{scheduler}, indicating the completion of a layer. Further details can be found in \cite{octopuscheduler25}.

The previous paper focused on enabling the functionality of single layers, as it was aimed at building preprocessing stages within broader applications. In contrast, this paper extends the approach to support end-to-end flows suitable for standalone edge applications, enabling the execution of complete models without further interaction by the host beyond initial configuration. The following section describes necessary extensions and modifications.

\section{Multi-Layer Scheduling}

\subsection{Overview}
Compared to single-layer \textit{OctopuScheduler}, the scheduling of multiple layers on chip requires having a unified basic structure for all layers. For this, the \textit{scheduler} and \textit{workers} are not model-specific, but instead rely on standardized layer execution and storing layer- and model-related configuration and parameters in DRAM. To accommodate models that exceed the available SRAM in the \textit{worker PEs}, inputs and outputs of all layers are stored in DRAM as well.

\subsection{Configuration Structure and DRAM Organization}
\begin{figure}
    \centering
    \includegraphics[width=0.95\linewidth]{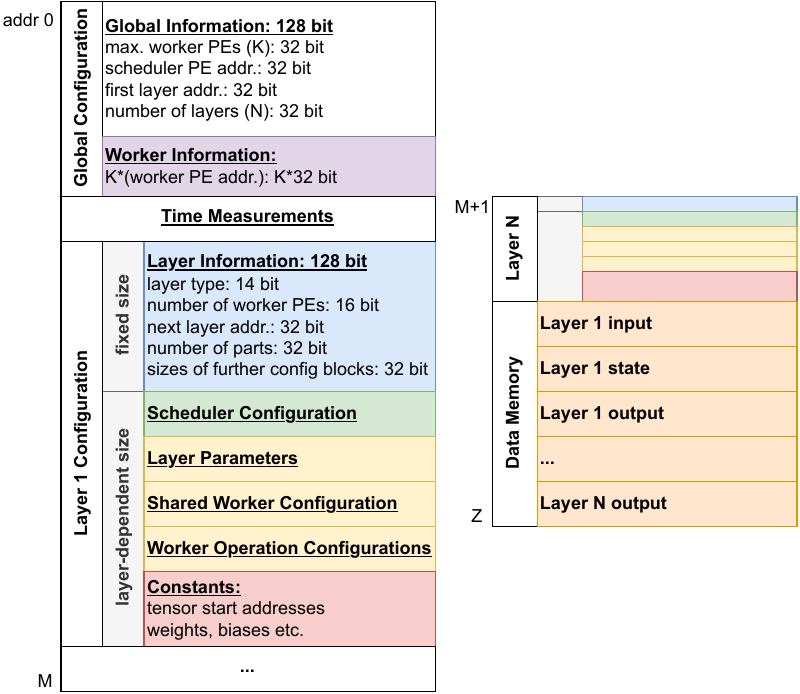}
    \caption{DRAM Memory Organization: Global Configuration is fixed at start of DRAM address space, followed by Layer Configurations and Data Memory. Continuous address space is split for compact visualization.}
    \label{fig:dram_structure}
\end{figure}

In order to avoid on-chip DRAM memory planning, the DRAM memory space is statically assigned by the host before execution.
It is organized in four subsequent main memory areas (see Fig. \ref{fig:dram_structure}):

\subsubsection{Global Configuration} 
This contains information necessary for the scheduling such as the coordinates of the \textit{scheduler} and the \textit{workers} and the number of layers in the model.

\subsubsection{Time Measurements} An area of memory is reserved for storing per-layer time measurements for each \textit{scheduler} and \textit{worker}, allowing for detailed profiling of model execution.

\subsubsection{Layer Configurations}
To enable efficient execution of diverse neural network operations across multiple \textit{workers} and allow for streamlined layer traversal, each layer is associated with a structured configuration block. This begins with a fixed-size header—referred to as the layer information, which encodes metadata such as the layer type (e.g., linear, add, softmax), the number of worker PEs assigned, and the DRAM address pointing to the configuration of the subsequent layer.

Beyond the header, the configuration includes several layer-specific sub-blocks with scheduler-specific and worker-specific configuration designed to optimize flexibility and reuse.

A subsequent section provides DRAM address mappings for input and output tiles per PE, enabling precise memory access during runtime, and any layer-specific constants, such as weights or biases. This organization not only maintains compactness in memory representation but also supports scalability across various layer types and topologies.

\subsubsection{Data Memory}
The final section in DRAM is used for the activation data, i.e. inputs, intermediate values and outputs of all layers in the model.

\subsection{On-Chip Model Execution}\label{subsec:on-chip-execution}
\begin{figure}
    \centering
    \includegraphics[width=0.85\linewidth]{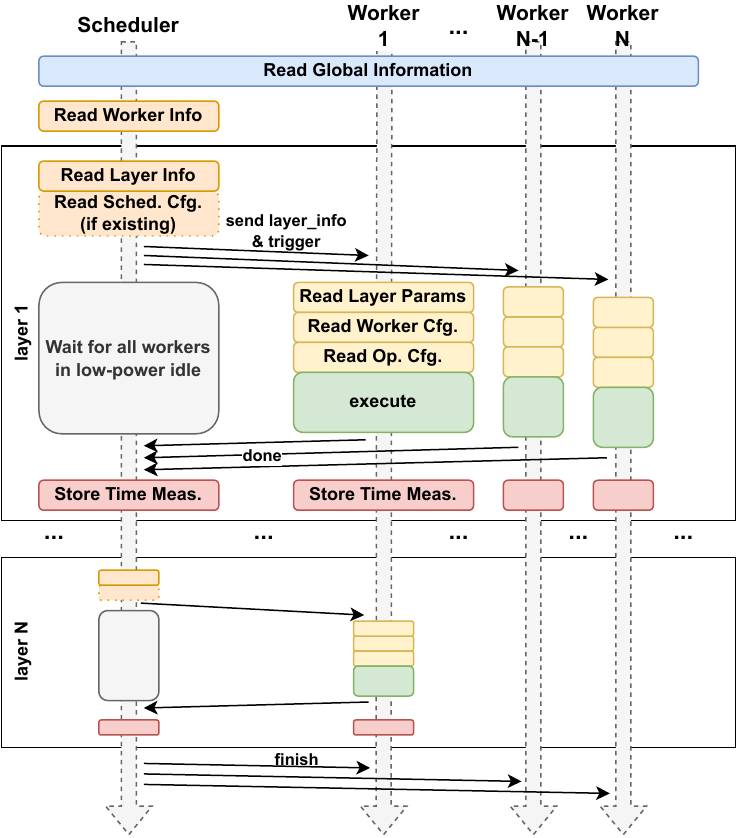}
    \caption{Scheduling flow}
    \label{fig:scheduling}
\end{figure}

The unified execution of models is visualized in Fig.~\ref{fig:octopuscheduler_overview} and~\ref{fig:scheduling}.
On startup, the \textit{scheduler} and all \textit{workers} start by reading the global information from DRAM. The \textit{scheduler} additionally reads the \textit{worker} information while the \textit{workers} go into low-power idle mode.

To execute the first layer, the \textit{scheduler} loads the layer information of the first layer from DRAM. From this, it knows which \textit{workers} to activate and the address of the next layer. It forwards the layer information and sends an interrupt to trigger all workers needed for executing the current layer. Each activated \textit{worker} individually fetches its further configuration data from DRAM and executes the layer-specific operations, potentially iterating over multiple parts. After successful completion, the \textit{workers} indicate the termination of their layer computations to the \textit{scheduler} and return to sleep mode.

When the \textit{scheduler} receives the completion flags from all \textit{workers}, it steps on to the next layer, until all layers have been executed. After the final layer, a special "finish" layer information is sent to all \textit{workers} to terminate their operation.

\section{Full Model Flow}
\subsection{Model Preparation}
\subsubsection{Quantization}
The end-to-end flow begins with a pre-trained PyTorch model, which is converted to an ONNX intermediate representation (IR). Post-training quantization from the AMD Quark library \cite{amdquark_2025} with INT8 power-of-two quantization is applied to the ONNX model for efficient hardware acceleration of the matrix multiplications on SpiNNaker2 using the MLA. To minimize quantization errors, Quark uses mean squared error optimization and cross-layer equalization \cite{nagel_data-free_2019} on a small calibration dataset.

\subsubsection{Lowering \& Layer Fusion}
The quantized ONNX graph is lowered into an application graph, a simplified IR containing the layers with parameters and quantization scales in the nodes and connections in the edges. This simplifies further graph manipulations, such as layer fusion.
Linear layer nodes are fused with immediately following ReLU nodes to exploit the ability of the MLA to execute matrix multiplications and ReLU as a joined operation.
Furthermore, unnecessary Quantize and Dequantize layers inserted by Quark (e.g. around softmax) are removed, pulling the quantization scaling factors into the related node.

\subsubsection{Layer Partitioning + Mapping}
To ensure that data dependencies of all layers are satisfied, the application graph is then topologically sorted into a linear chain of nodes in the order of planned execution on hardware.

The sorted nodes are converted into our low-level \textit{S2Layer} representation. In addition to implementing the partitioning and mapping mechanisms from \textit{Octopuscheduler} \cite{octopuscheduler25}, this class library defines memory allocation and generation of layer configurations, as well as reference models for each layer type. 
The \textit{S2Layers} of the model are wrapped into an \textit{S2Model}, which assembles the complete DRAM structure based on the individual layers, including configurations and constants and plans layer input and output locations.

\subsection{Model Execution}
After model preparation, the model can be executed (see Fig.~\ref{fig:octopuscheduler_overview}). First, the configuration and constants part of the DRAM structure is written into DRAM by the host.

Next, the host writes the input data to DRAM and notifies the scheduler via an interrupt, at which point the on-chip scheduling per Section~\ref{subsec:on-chip-execution} begins.
After the computation of the model has terminated, the outputs of the neural network can be read out by the host.

\section{Experimental Results}

\subsection{Dummy Model for Testing the End-to-End Flow}
\begin{table}[tb]
\centering
\caption{Accuracy (\%) on MNIST Validation Set for Quantized MLP}
\label{tab:resnet_quant}
\begin{tabular}{lcc}
\toprule
 & accuracy \\
\midrule
pre-trained FP32 & 98.36 \\
\midrule
INT8 Power-of-Two MSE CLE & 98.33  \\
Measured on SpiNNaker2 & 98.34 \\
\bottomrule
\end{tabular}
\end{table}
For verifying our implementation, we implement and train a simple three-layer MLP in PyTorch on MNIST with 512 and 256 neurons for the hidden layers. To accomodate the minimum tile size for efficient use of the MLA, the output size is padded from 10 to 16. The model trained in 32-bit floating-point precision in PyTorch achieves 98.36\,\% accuracy on the validation set, which is reduced to 98.33\,\% after quantization in ONNX. Running the quantized model on SpiNNaker2 using our pipeline yields an accuracy of 98.34\,\%. The negligible differences are attributed to the fact that the quantized ONNX model is not a bit-accurate model of execution on SpiNNaker2.

\subsection{Runtime Analysis}
\begin{table}
\footnotesize
\centering
\caption{Runtime breakdown for a 784-512-256-16 MLP on SpiNNaker2.} 
\label{tab:per_layer_runtime}
\begin{tabular}{lcccc}
\toprule
 & Runtime  & Input   & Output & Workers \\
\midrule
Setup$^{*}$ &  12\,µs &      &   \\
\midrule
FC1+ReLU    & 323\,µs & 784 &  512 & 8  \\
FC2+ReLU    & 217\,µs & 512 &  256 & 4  \\
FC3         &  80\,µs & 256 &   16 & 1  \\ 
Softmax     &  50\,µs &  16 &   16 & 1  \\
\midrule
Cleanup$^\dagger$ & 9\,µs &     &      &    \\
\midrule
Total       & 688\,µs &     & \\
\bottomrule
\end{tabular}
\scriptsize

$^{*}$ Time to configure all \textit{worker} addresses and \textit{workers} in \textit{scheduler} at startup. \\
$^\dagger$ Time to shut down \textit{workers} and \textit{scheduler} and store time measurements in DRAM. \\
\end{table}

To analyze the runtime of the model, we measure the execution time on the SpiNNaker2 chip for batch size 1. 
As the duration of host interaction apart from writing inputs and reading results is negligible, it is excluded from time measurements to show pure execution times.
The total runtime is 688\,µs (Table~\ref{tab:per_layer_runtime}), with most time spent in the larger first two linear layers, contributing 323\,µs and 217\,µs, respectively.

The execution time of the linear layers is dominated by loading the weights from DRAM. In the first layer, fetching the weights takes on average 192\,µs out of the total layer duration of 323\,µs, whereas the actual computation of the vector-matrix multiplication on the MLA only needs 29\,µs. This means that the accelerator is only used for 9\,\% of the layer runtime.

\subsection{Scheduling Overhead}
The model overhead for setup (before execution) and cleanup (after execution) is small at 12\,µs and 9\,µs as only 8 \textit{workers} are enabled. If all available \textit{workers} on a single chip are enabled, this increases to 39\,µs and 93\,µs, which is still only $\sim$16\,\% of the total runtime of this small model.

Each layer incurs on average 13\,µs of overhead for the scheduling. Whereas this is negligible for the first two linear layers, it becomes a more significant fraction of the runtime of the small final linear layer and the small softmax layer. This overhead mainly consists of loading the layer information from DRAM, distributing it and triggering the relevant \textit{workers}, as well as storing the time measurements of the layer.

\section{Discussion}
In this work, we present an end-to-end DNN inference framework for the neuromorphic SpiNNaker2 platform crafted towards edge applications as an extension of the previous \textit{OctopuScheduler} framework. In a fully automated flow, it takes a DNN model trained in PyTorch, applies post-training quantization and executes it on a SpiNNaker2 chip, making usage of the available accelerators. Previous works still relied on either manual model-specific implementations  \cite{nazeer_language_2024} or were limited to single layers \cite{octopuscheduler25}.

Our framework is geared towards single-chip edge applications with models that exceed the available SRAM of a single SpiNNaker2 chip.
Hence, the dummy MLP model presented in this paper demonstrates only preliminary results of our proposed method, but already shows very low scheduling overhead.
Larger models, such as ResNet architectures or small language models, will fully exploit the parallelism of SpiNNaker2 and unlock the full potential of our approach.

To achieve highest throughput, approaches that avoid DRAM completely and only use PE SRAM should be considered, but are only feasible if the model fits into the available PE memory. Otherwise, hiding the DRAM latency behind computation should be investigated. Alternatively, the model inference could be extended over multiple SpiNNaker2 chips.

Future work will include further optimizations, particularly aimed at reducing and accelerating the data transfers between SpiNNaker2 and DRAM, which currently represent the main bottleneck. Future work will also involve integrating the edge-based approach presented here into the proprietary stack being developed at SpiNNcloud for large-scale systems.

\section*{Acknowledgment}
We would like to thank Jiaxin Huang and the team at SpiNNcloud Systems GmbH for helpful technical discussions.

\bibliographystyle{IEEEtran}
\bibliography{main.bib}

\end{document}